\documentclass[letterpaper, 10 pt, conference]{ieeeconf}

\IEEEoverridecommandlockouts
\usepackage{amsmath,amssymb,amsfonts,siunitx}
\usepackage{graphicx,epsfig,layouts}

\let\proof\relax
\usepackage{amsthm}

\let\labelindent\relax
\usepackage{enumitem}
\usepackage{comment}
\usepackage{tabularx}

\usepackage[dvipsnames]{xcolor}
\usepackage[font=small,labelfont=bf]{caption}
\usepackage[position=b]{subcaption}
\usepackage[utf8]{inputenc}

\usepackage{flushend}   

\newcommand{\Alpha}{\textit{Alpha}}
\newcommand{\Bravo}{\textit{Bravo}}
\newcommand{\Al}{\mathcal{A}}
\newcommand{\Br}{\mathcal{B}}

\newcommand{\vectorvar}[1]{{\boldsymbol{\mathrm{#1}}}}
\newcommand{\bsym}[1]{\boldsymbol{#1}}
\newcommand{\iden}[2]{\mathbb{I}_{#1\times#2}}
\newcommand{\zero}[2]{0_{#1\times#2}}
\newcommand{\SO}[1]{\mathsf{SO}(#1)}
\newcommand{\R}[1]{\mathbb{R}^{#1}}
\newcommand{\proj}[1]{\textnormal{proj}_{#1}}
\newcommand{\fext}{\vectorvar{f}_{\mathrm{ext}}}

\newtheorem{definition}{Definition}
\newtheorem{assumption}{Assumption}
\newtheorem{theorem}{Theorem}

\usepackage{soul}
\usepackage[most]{tcolorbox}  

\title{\LARGE \bf SO(2)-Equivariant Downwash Models for Close Proximity Flight}
\author{Henry Smith$^*$, Ajay Shankar, Jennifer Gielis, Jan Blumenkamp, and Amanda Prorok
\thanks{This work was supported by ARL DCIST CRA W911NF-17-2-0181, European Research Council (ERC) Project 949940 (gAIa), and, in part, by a gift from Arm.} 
\thanks{Authors are with the Department of Computer Science \& Technology, University of Cambridge, UK. Emails: \texttt{\{hds35, as3233, jag233, jb2270, asp45\}@cl.cam.ac.uk}}
\thanks{*Author for correspondence.}
}

\begin{document}
\maketitle
\thispagestyle{empty}
\pagestyle{empty}
\begin{abstract}
Multirotors flying in close proximity induce aerodynamic wake effects on each other through propeller downwash.
Conventional methods have fallen short of providing adequate 3D force-based models that can be incorporated into robust control paradigms for deploying dense formations.
Thus, \textit{learning} a model for these downwash patterns presents an attractive solution. 
In this paper, we present a novel learning-based approach for modelling the downwash forces that exploits the latent geometries (i.e.~symmetries) present in the problem. 
We demonstrate that when trained with only \SI{5}{} minutes of real-world flight data, our geometry-aware model outperforms state-of-the-art baseline models trained with more than \SI{15}{} minutes of data.
In dense real-world flights with two vehicles, deploying our model online improves 3D trajectory tracking by nearly \SI{36}{\percent} on average (and vertical tracking by \SI{56}{\percent}).
\end{abstract}


\section{Introduction}\label{sec:introduction}
Multi-robot tasks often require aerial robots to fly in close proximity to each other.
Such situations occur during collaborative mapping and exploration missions, which may require the robots to navigate constricted areas \cite{preiss2017crazyswarm,vasarhelyi2018optimized}, or when the task is constrained in a more limited workspace from the outset (ex.~indoors) \cite{turpin2012trajectory}.
In some cases, such as aerial docking and payload transport~\cite{shankar2020dynamic, miyazaki2018airborne, shankar2021multirotor}, a close approach to another multirotor is indeed intended.
The aerodynamic interference from other vehicles in all these cases is an additional risk and constraint for motion planners.

While it is possible to extract computational fluid models for multirotors that capture aerodynamic interactions over the entire state-space of the problem, such high-fidelity models \cite{yoon2017computational,yoon2016computational} are often too expensive and restrictive (computational time and run-time memory), or simply unnecessary (dynamically transitioning flight modes).
To enable complex and fluid flight missions, we require a fast and accurate model of these exogenic forces that can facilitate onboard controllers robust to these disturbances.

In this work, we present a novel learning-based approach for estimating the downwash forces produced by a single multirotor. 
Unlike previous learning-based approaches, our \textit{equivariant downwash model} makes assumptions on the geometry present in the underlying downwash function. 
To encode these assumptions in our model, we extract \textit{invariant geometric features} from the input data, which decreases the dimensionality of the learning problem. 
Whereas traditional machine learning algorithms often require large amounts of training data to accurately learn the underlying function \cite{shalev2014understanding}, our geometry-aware algorithm is sample-efficient. 

We train the equivariant downwash model on real-world flight data collected by two multirotors using a baseline model-based controller (a linear quadratic regulator, LQR).
When deployed online within the controller, our model achieves state-of-the-art (SOTA) performance on a variety of challenging experiments.
To our knowledge, the equivariant downwash model is the first learning-based approach to uncover consistent patterns in the downwash in the lateral plane.
Further, we empirically validate that our model is more sample-efficient than SOTA learning-based approaches.

\begin{figure}[t]
    \centering
    \includegraphics[width=0.95\linewidth,trim={0 6cm 0 0},clip]
        {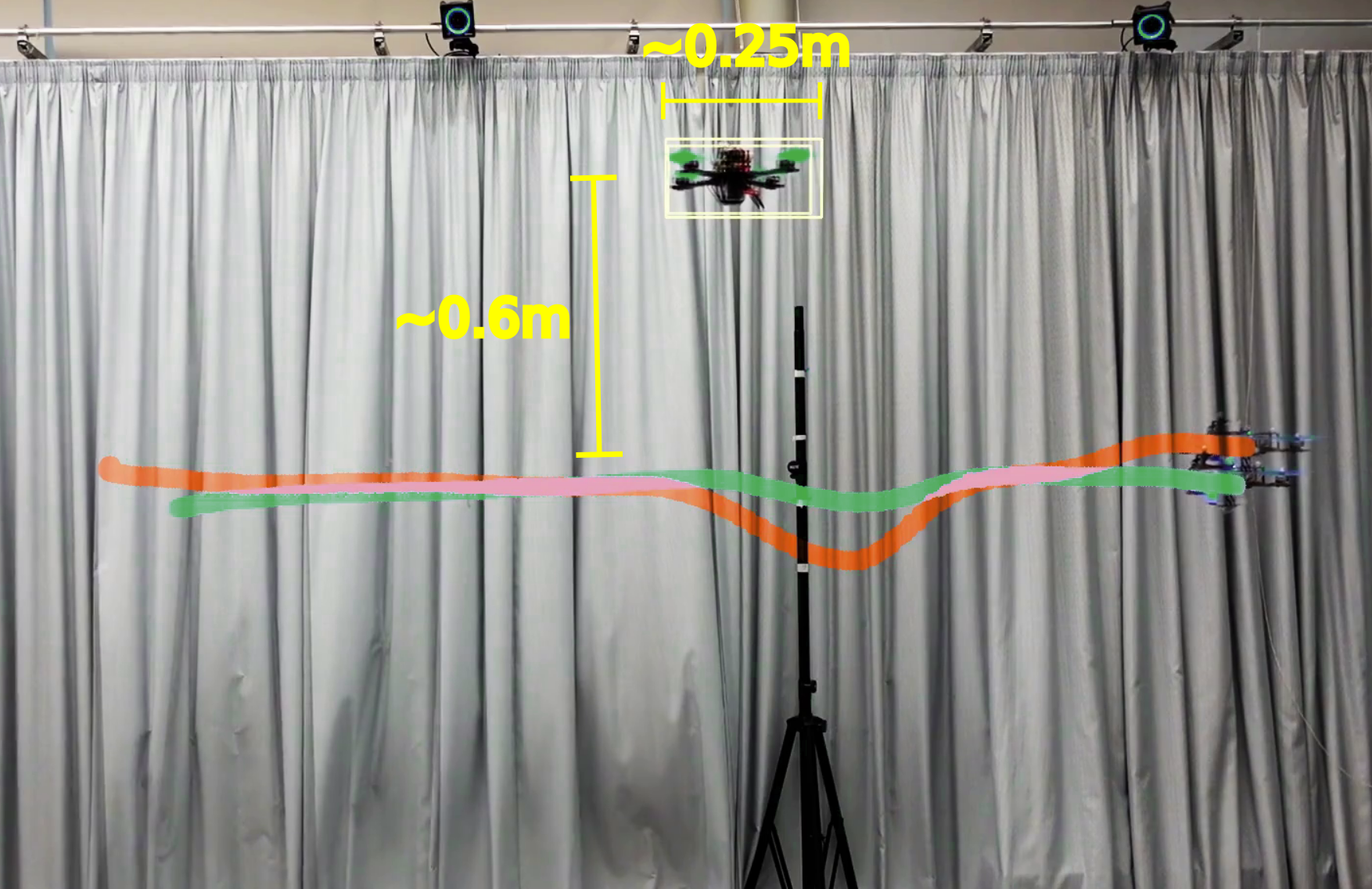}
    \caption{A snapshot demonstrating the improvement in trajectory tracking with (green) and without (red) our downwash model.}
    \vskip-2.5ex
    \label{fig:cover_image}
\end{figure}

\subsection{Related Work}
\textbf{Multi-UAS Flights.}
Despite recent and increasing interest in the problem of tight formations in aerial swarms~\cite{chung2018}, there is a dearth of work that attempts to \textit{deploy} teams in close proximity.
Downwash-induced forces, and the resulting deviations from planned trajectories, can be trivially minimized by enforcing hard inter-agent separation constraints that are large enough for simplified motion planners~\cite{zhou2021egoswarm}.
Such approaches severely limit achievable swarm density by na\"ively excluding navigable airspace, and worse, may fail their intended mission objectives due to the chaotic and directional nature of both single- and multi-agent aerodynamic interactions \cite{throneberry2021, zhang2020numerical}.

Recent work has combined physics-based nominal dynamics models with deep neural networks (DNN) to predict and counteract exogenous forces from ground effect \cite{shi2019neural, shi2021neural2} or from neighboring multirotors \cite{shi2020neural, shi2021neural2, li2023nonlinear}.
Modeling the discrete-time (residual) dynamics of a multirotor has also been studied using, for instance, Gaussian processes (GPs) \cite{wang2018safe} and recurrent neural networks (RNNs) \cite{mohajerin2018deep}.
Others have directly learned the continuous-time dynamics using neural ordinary differential equations (neural-ODEs) combined with model predictive  controllers~\cite{chen2018neural,chee2022knode}.
However, extending these works to model inter-vehicle
interaction dynamics can be non-trivial. 
This is because in this work, we are representing the geometric symmetries in the problem, and modeling their evolution adds sensing and computational complexities.
We thus consider the model from \cite{shi2019neural} the SOTA for learning-based downwash prediction.

\textbf{Geometric Deep Learning.}
At its core, geometric deep learning involves imposing inductive biases (called ``geometric priors") on the learning algorithm via one's knowledge of the problem geometry.
As we will discuss in Section \ref{sec:geometriclearning}, these geometric priors are represented using assumptions of invariance and equivariance on the underlying function being learned \cite{bronstein2021geometric,esteves2020theoretical}.
The purpose of geometric priors is that they intuitively correspond to ``parameter sharing," and thus have been shown to improve sample efficiency across various applications \cite{wang2023surprising, wang2022equivariant}. 

Within the field of robotics, equivariant reinforcement learning has been employed for low-level quadrotor control \cite{yu2023equivariant} and robot manipulation tasks \cite{zhu2022grasp, wang2022robot}. 
In particular, recent research has developed equivariant deep-$Q$ learning and soft actor-critic algorithms \cite{wang2022equivariant} capable of learning complex manipulation tasks (ex.~grasping, picking up and pushing blocks) using only a couple of hours of training data \cite{zhu2022grasp, wang2022robot}. 
These learning algorithms were performed entirely on physical robots (i.e.~on-robot learning). 
To our knowledge, there has been no previous work that utilizes geometric priors to efficiently learn multirotor downwash forces.

\subsection{Contributions} 
The key contributions of our work are as follows:
\begin{enumerate}
    \item We propose an equivariant model for multirotor downwash that makes assumptions on the downwash field geometry. 
    This geometry-aware model represents data in a lower-dimensional space in order to satisfy the assumed rotational equivariance of our system.
    \item We provide real-world experimental results that showcase the sample efficiency of our equivariant downwash model. 
    Using only $5$ minutes of flight data, we learn the downwash function with greater accuracy than SOTA learning-based approaches do with $15$ minutes of data.
    \item When deployed online within an optimal feedback controller, our model's predictions reduce vertical tracking errors by \SI{56}{\percent} and lateral tracking errors by \SI{36}{\percent}.
\end{enumerate}
\section{Problem Formulation}
Throughout the paper, we consider two identical multirotor vehicles, referred to as \Alpha{} ($\mathcal{A}$) and \Bravo{} ($\mathcal{B}$), operating in close proximity of one another.
They have similar estimation and control stacks onboard, with the only difference being in their reference states/trajectories as well as the additional force correction terms.
We will assume \Alpha{} is a ``leader" aircraft, while \Bravo{} is a ``follower" that suffers under the propeller downwash generated by \Alpha{}.

\textbf{Notation.}
We use $\SO{n}$ to refer to the special orthogonal group. The elements in $\SO{2}$ and $\SO{3}$ represent two- and three-dimensional rotations about a point and a line, respectively.
Unless explicitly specified, we will assume that all frames follow the North-East-Down (NED) convention with a right-hand chirality.
We will let $\Al = \{a_1, a_2, a_3\}$ denote the body frame of \Alpha{} and $\mathcal{M} = \{\hat{e}_1, \hat{e}_2, \hat{e}_3\}$ denote the inertial frame (and the corresponding unit vectors).
The matrix $R_{\mathcal{M}}^{\mathcal{C}} \in \SO{3}$ denotes the rotation matrix from the inertial frame to a body frame $\mathcal{C}$ (which is a unitary transformation, i.e., $(R_{\mathcal{M}}^{\mathcal{C}})^\top = (R_{\mathcal{M}}^{\mathcal{C}})^{-1}$).

Additionally, we will use the notation $\vectorvar{x}$ for vectors, and $\bsym{x}$ for vector-valued functions of time. 
For an $n$-dimensional vector $\vectorvar{x}$, we will let $[\vectorvar{x}]_i$  denote its $i$th component. 
All vectors are assumed to be in the inertial frame, $\mathcal{M}$. The position and velocity vectors corresponding to \Alpha{} and \Bravo{} will be written with the superscripts $\Al$ and $\Br$, respectively. 
We will abbreviate sine and cosine functions as $s(\cdot)$ and $c(\cdot)$.

Lastly, for the frame of the ``leader," \Alpha{}, we define the subspaces $\mathcal{S}_{\Al} \equiv \text{span}\{a_1, a_2\}$ and $\mathcal{S}_{\Al}^{\perp} = \text{span}\{a_3\}$. 
The  operator $\proj{\mathcal{S}}: \mathbb{R}^3 \rightarrow \mathbb{R}^3$ maps each vector in $\mathbb{R}^3$ to its orthogonal projection in subspace $\mathcal{S}$.

\textbf{Multirotor Dynamics/Control.}
We model a multirotor as a rigid body $\mathcal{C}$ with six degrees of freedom with mass \textit{m}, and dynamics in an inertial NED frame governed by
\begin{equation}
  m \vectorvar{a} = -R_{\mathcal{M}}^{\mathcal{C}} T + \hat{e}_{3} mg,
  \label{eq:rigid-body}
\end{equation}
where $T$ is the collective thrust produced by the rotors and $g$ is the acceleration due to gravity.
The matrix $R_{\mathcal{M}}^{\mathcal{C}}$ is composed from the Euler roll ($\phi$), pitch ($\theta$) and yaw ($\psi$) angles of the body in Z-Y-X rotation order.
A nominal controller for this system generates the control targets
$\bsym{u} = [\phi, \theta, \dot{\psi}, T]^\top$ using a non-linear inversion map on \eqref{eq:rigid-body} to affect a desired acceleration, $\vectorvar{a} \in \R{3}$.
This allows us to write the system of equations in a linear form,
\begin{equation}
  \dot{\bsym{x}} = A\bsym{x} + B\bsym{u},
  \text{ and, }
  y = C\boldsymbol{x}
  \label{eq:rigid-body-lin}
\end{equation}
with
$\bsym{x}(t) = [p_n, p_e, p_d, v_n, v_e, v_d, \psi]^\top
\equiv
[\bsym{p}, \bsym{v}, \psi]^\top$
representing the state vector,
$\bsym{u}(t) = [a_n, a_e, a_d, \dot{\psi}]^\top \equiv [\vectorvar{a}, \dot{\psi}]^\top$ the feedback-linearized control input, and
{\small
\begin{equation} 
  \begingroup
  \setlength\arraycolsep{3pt}
  A = \begin{pmatrix}
        \zero{3}{3} & \iden{3}{3} & \zero{3}{1} \\
        \zero{3}{3} & \zero{3}{3} & \zero{3}{1} \\
        \zero{1}{3} & \zero{1}{3} & 0 \\
      \end{pmatrix}, 
  B = \begin{pmatrix}
        \zero{3}{3} & \zero{3}{1} \\
        \iden{3}{3} & \zero{3}{1} \\
        \zero{1}{3} & 1
      \end{pmatrix},
  C = \mathbb{I}_{7 \times 7}. \nonumber
  \endgroup
\end{equation}
}

Since $(A,B)$ is controllable, it is then straightforward to derive an optimal stabilizing control law, $\bsym{u}(t) = -K(\bsym{x}(t)-\bsym{x_r}(t))$, that regulates this second-order plant to a reference state $\bsym{x_r}$.
The gain matrix, $K$, is designed with a linear quadratic regulator (LQR) to produce a high gain margin.

\textbf{Downwash Model.}
In this work, we model the aerodynamic downwash effects, $\fext \in \mathbb{R}^3$, experienced by a multirotor as additive exogenic forces (or equivalently, accelerations) acting on \eqref{eq:rigid-body-lin}.
We assume that these forces can be written as $\fext \equiv \fext(\mathbf{x})$, where $\mathbf{x} = [\mathbf{p}^{\Al}, \mathbf{p}^{\Br}, \mathbf{v}^{\Al}, \mathbf{v}^{\Br}]$ contains the instantaneous state information of \Alpha{} and \Bravo{}.
The second-order model described above abstracts the torques produced by per-motor thrust differentials and delegates the regulation of angular states to a well-tuned low-level autopilot.
This method of successive loop-closure \cite{gorder1997sequential} allows us to model the short-term torque dynamics induced by aerodynamic interactions as collective forces, thereby generalizing the method to other types of aircraft.
Hence, \eqref{eq:rigid-body-lin} can now be rewritten as
$\dot{\bsym{x}} = A\bsym{x} + B\bsym{u} + B\fext = A\bsym{x} + B(\bsym{u} + \fext)$.
Since our control, $\bsym{u}$, is designed with very high gain margins, we can use this linear separability to adapt the feedback control to compensate for this effect as
$\bsym{u}_\vectorvar{f}(t) \equiv -K(\bsym{x}(t)-\bsym{x_r}(t)) - \fext$.
We note that, in the general case where the controller's stability margin may be narrow, this compensation can be incorporated through constraint-based methods (e.g.~model predictive control).

\textbf{Problem.}
Our objective is to learn a sample-efficient model that predicts $\fext(\mathbf{x})$ such that a closed-loop controller can compensate for predicted exogenic disturbances online.
\section{Establishing Geometric Priors}\label{sec:geometriclearning}
In order to efficiently and accurately model the downwash forces experienced by \Bravo{}, we first make assumptions on the geometry present in $\fext(\mathbf{x})$. 
Our assumptions are formalized using the group-theoretic definitions of invariance and equivariance.

\subsection{Geometric Invariance and Equivariance}

\begin{figure}
    \centering
    \includegraphics[width=0.48\textwidth]{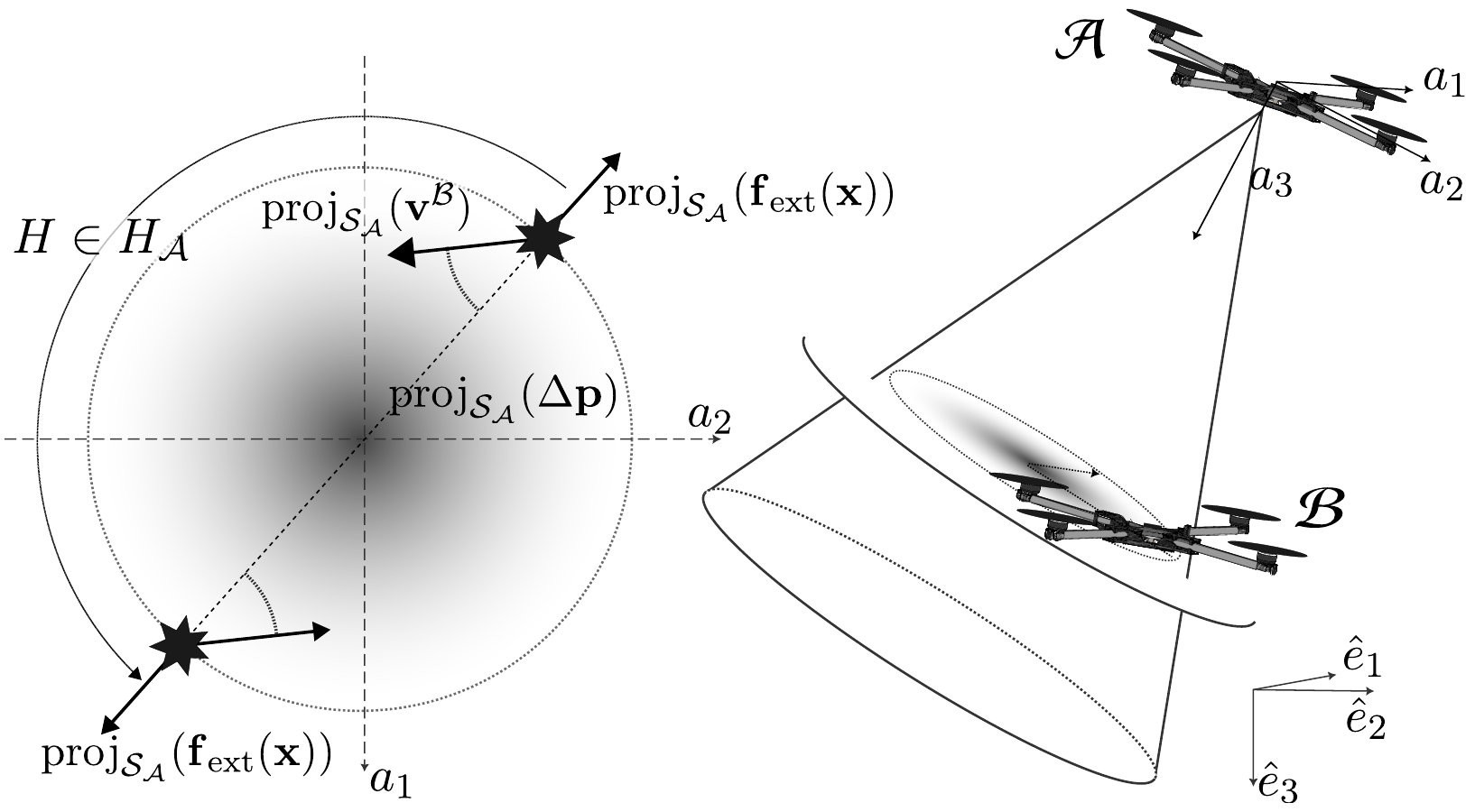}
    \caption{An illustration of Assumption \ref{assumption:equivariance} on the downwash function $\fext(\mathbf{x})$. 
    On the left, we provide two combinations of $(\Delta \mathbf{p}, \mathbf{v}^{\Br})$ that are related under the rotational equivariance property.}
    \label{fig:frames-notation}
\end{figure}

The geometric properties of functions are described in terms of group actions.
To be specific, let $G$ be a group and $\mathcal{X}$ be a set. The \textit{action} of group $G$ on set $\mathcal{X}$ is a mapping $\star: G \times \mathcal{X} \rightarrow \mathcal{X}$ which associates with each group element and set element a corresponding set element.
The group action $\star$ must satisfy certain properties \cite{bronstein2021geometric,esteves2020theoretical}. 
In this case, we say that ``$G$ acts on $\mathcal{X}$ according to $\star$."
    
For instance, if $G = \SO{2}$ and $\mathcal{X} = \mathbb{R}^2$, then $G$ can act on $\mathcal{X}$ according to matrix multiplication: $G_{\omega} \star \mathbf{w} = G_{\omega} \mathbf{w}$, where $G_{\omega} \in \SO{2}$ is the rotation matrix corresponding to the angle $\omega \in [0, 2\pi)$ and $\mathbf{w} \in \mathbb{R}^2$ is an arbitrary vector.\footnote{This group action is technically a group representation: an action of $G$ on a vector space by invertible linear transformations.}

Using group actions, one can define \textit{invariant} and \textit{equivariant} functions:
\begin{definition}[Invariance, Equivariance]\label{def:equivariance}
Let $G$ be a group and $\mathcal{X}, \mathcal{Y}$ be two sets. 
Suppose that $G$ acts on $\mathcal{X}$ according to $\star_1$ and on $\mathcal{Y}$ according to $\star_2$. 
    
A function $f: \mathcal{X} \rightarrow \mathcal{Y}$ is \textbf{invariant} with respect to $\star_1$ if it satisfies
\begin{align*}
    f(x) = f(g \star_1 x), \quad \forall x \in \mathcal{X}, \forall g \in G.
\end{align*}
$f$ is \textbf{equivariant} with respect to $\star_1$ and $\star_2$ if it satisfies
\begin{align*}
    g \star_2 f(x) = f(g \star_1 x), \quad \forall x \in \mathcal{X}, \forall g \in G.
\end{align*}
\end{definition}
Intuitively, invariance states that the output of $f$ should be preserved regardless of whether or not $g \in G$ acts on the input.
Equivariance, on the other hand, states that $g \in G$ acting on the input $x$ according to $\star_1$ is equivalent to $g$ acting on the output of $f$, $f(x)$, according to $\star_2$.

\subsection{Geometric Assumptions on $\fext$}\label{sec:geomlearn}

Now that we have detailed the geometric properties a function may have, we consider the particular structure of the interaction forces $\mathbf{f}_{\text{ext}}(\mathbf{x})$ that \Alpha{} exerts on \Bravo{}.

Foremost, we know that $\fext(\mathbf{x})$ should not depend on positional shifts in the input space:
\begin{assumption}[Translation Invariance]\label{assumption:invariance}
Define the group $\mathbb{T}$ consisting of all translations in $\mathbb{R}^3$. $\mathbb{T}$ is isomorphic to $\mathbb{R}^3$. 
We assume $\fext(\mathbf{x})$ is invariant with respect to the group action $\mathbf{t} \star_1 \mathbf{x} = [\mathbf{t} + \vectorvar{p}^{\Al}, \mathbf{t} + \vectorvar{p}^{\Br}, \vectorvar{v}^{\Al}, \vectorvar{v}^{\Br}]$ for $\mathbf{t} \in \mathbb{T}$. 
\end{assumption}
Equivalently, Assumption \ref{assumption:invariance} states that $\mathbf{f}_{\text{ext}}$ must be a function of $\Delta \vectorvar{p}$, $\vectorvar{v}^{\Al}$, and $\vectorvar{v}^{\Br}$ only. 
From here on, we will redefine $\mathbf{x} \equiv [\Delta \vectorvar{p}$, $\vectorvar{v}^{\Al}$, $\vectorvar{v}^{\Br}] \in \mathbb{R}^9$. 
Translation invariance is standard in downwash models \cite{shi2020neural, shi2021neural2, li2023nonlinear}. 
    
However, beyond translation invariance, previous downwash models have failed to consider the geometry present in $\fext$.
In particular, in many flight regimes, it is reasonable to assume that once the downward direction $a_3$ of the ``leader" \Alpha{} is fixed, how one defines the north and east directions is arbitrary. 
This is the subject of the following assumption:
\begin{assumption}[Rotational Equivariance]\label{assumption:equivariance}
Define the group $H_{\Al} \subset \SO{3}$ containing all rotations that fix $a_3$, the down direction in the body frame of $\Alpha$:
\begin{equation}\label{group}
         H_{\Al} \! = \! \big\{ H \in \SO{3} \ | \ \textnormal{proj}_{\mathcal{S}_{\Al}^{\perp}} \left(H\mathbf{w}\right) \! = \! \textnormal{proj}_{\mathcal{S}_{\Al}^{\perp}}\left(\mathbf{w}\right)\! , \! \forall \mathbf{w} \in \mathbb{R}^3 \big\}.
\end{equation}
$H_{\Al}$ is isomorphic to the two-dimensional rotation group, $\SO{2}$. 
Define the action of $H_{\Al}$ on the input space by $H \star_1 \mathbf{x} = [H\Delta \mathbf{p} , \mathbf{v}^{\Al}, H\mathbf{v}^{\Br}]$ and on the output space by $H \star_2 \mathbf{w} = H\mathbf{w}, \mathbf{w} \in \mathbb{R}^3$ for $H \in H_{\Al}$. 
Then we assume $\fext = \fext(\mathbf{x})$ is equivariant with respect to these group actions.
\end{assumption}

This rotational equivariance assumption is illustrated in Figure \ref{fig:frames-notation}. 
Intuitively, Assumption \ref{assumption:equivariance} states that in the frame of the leader vehicle \Alpha{}, rotating the relative position vector and the velocity vector of \Bravo{} in the $\{a_1, a_2\}$ axes by an angle of $\omega \in [0, 2\pi)$ is equivalent to rotating the force vector $\mathbf{f}_{\text{ext}}(\mathbf{x})$ of \Bravo{} by the same angle $\omega$.

We clarify that the true downwash function will not necessarily satisfy Assumption \ref{assumption:equivariance} in all cases.
For instance, it is unlikely to hold when \Alpha's rotor speeds are highly asymmetric (ex.~during aggressive maneuvering)~\cite{zhang2020numerical}.
However, as we will demonstrate in Section \ref{sec:experiments}, imposing this geometric prior on the learning algorithm results in significant improvements in sample efficiency without incurring a large bias.
This result is in line with recent research \cite{wang2023surprising} suggesting that even when the assumed geometric priors do not exactly match the underlying symmetry (i.e.~``approximate" equivariances), they can still yield gains in sample efficiency while outperforming non-equivariant models.

Also, we note that, when Assumption \ref{assumption:equivariance} is not strictly satisfied by the underlying system, there still exist symmetries in the aerodynamic forces that can and should be exploited. 
In particular, when the entire system is rotated by $H \in H_{\mathcal{S}_{\{\hat{e}_1, \hat{e}_2 \}}}$, a rotation which fixes the down axis $\hat{e}_3$ of the inertial frame, the forces will rotate by the same $H$. That is to say, although the forces may no longer be rotationally equivariant with respect to the down axis of the leader $a_3$, they will be about the down axis of the inertial frame.\\
\indent The model that we propose in Section \ref{sec:geomlearning} allows one to account for this equivariance in the inertial frame \textit{without} imposing Assumption \ref{assumption:equivariance}. In particular, one can take the subspace $\mathcal{S}_{\Al}$ to be $\mathcal{S}_{\{\hat{e}_1, \hat{e}_2 \}}$ and the change-of-basis transformation $R_{\mathcal{M}}^{\Al}$ to be the identity.
\section{Geometry-Aware Learning}\label{sec:geomlearning}
Now that we have stated our assumptions on $\fext(\mathbf{x})$, we encode them as geometric priors in our learning algorithm.

\subsection{Rotationally Equivariant Model}

In order to present our model for $\fext(\mathbf{x})$, we first define a feature mapping $h: \mathbb{R}^{9} \rightarrow \mathbb{R}^6$
\begin{align}\label{featuremap}
    &h(\mathbf{x}) = \bigg(\! \frac{ \proj{\mathcal{S}_{\Al}}(\Delta \vectorvar{p} )^{\top} \proj{\mathcal{S}_{\Al}}(\vectorvar{v}^{\Br} )}{\| \proj{\mathcal{S}_{\Al}}(\Delta \vectorvar{p})\|_2 \|\proj{\mathcal{S}_{\Al}}(\vectorvar{v}^{\Br} )\|_2}, \| \proj{\mathcal{S}_{\Al}}\left(\Delta \vectorvar{p} \right)\|_2,\nonumber\\
    & \| \proj{\mathcal{S}_{\Al}}(\vectorvar{v}^{\Br})\|_2, 
    \left[R_{\mathcal{M}}^{\Al} \Delta \vectorvar{p}\right]_3, 
    \left[R_{\mathcal{M}}^{\Al} \vectorvar{v}^{\Br} \right]_3, \| \proj{\mathcal{S}_{\Al}}(\vectorvar{v}^{\Al})\|_2 \!\bigg).
\end{align}

The mapping $\vectorvar{x} \mapsto h(\vectorvar{x})$ transforms each input vector $\vectorvar{x}$ in Euclidean space into an invariant representation with respect to the action of $H_{\Al}$.
It does so by separating each of the inputs $\Delta \mathbf{p}$,  $\mathbf{v}^{\Al}$, and $\mathbf{v}^{\Br}$ into their components in the subspaces $\mathcal{S}_{\Al}$ and $\mathcal{S}_{\Al}^{\perp}$. 

In particular, because the components of $\Delta \mathbf{p}$ and $\mathbf{v}^{\Br}$ contained in $\mathcal{S}_{\Al}^{\perp}$, $\proj{\mathcal{S}_{\Al}^{\perp}}(\Delta \mathbf{p})$ and $\proj{\mathcal{S}_{\Al}^{\perp}}(\mathbf{v}^{\Br})$, are unaffected by the action of $H_{\Al}$, then our model has the freedom to operate on them arbitrarily. 
These components can be rewritten in the frame of \Alpha{} as $\left[R_{\mathcal{M}}^{\Al} \Delta \vectorvar{p}\right]_3$ and $\left[R_{\mathcal{M}}^{\Al} \vectorvar{v}^{\Br} \right]_3$. 
The components contained in $\mathcal{S}_{\Al}$, on the other hand, $\proj{\mathcal{S}_{\Al}}(\Delta \mathbf{p})$ and $\proj{\mathcal{S}_{\Al}}(\mathbf{v}^{\Br})$, are affected by the action $\star_1$ of $H_{\Al}$. 
Therefore, we only consider the magnitudes of these vectors as well as the angles between them.
Formal verifications of these statements are given in the proof of Theorem \ref{thm:equivariance}.

While the feature mapping \eqref{featuremap} we proposed is invariant with respect to the action of $H_{\Al}$, we ultimately want our model for $\vectorvar{f}_{\text{ext}}(\mathbf{x})$ to be equivariant with respect to $\star_1$ and $\star_2$.
We achieve this by taking into account the polar angle that $\proj{\mathcal{S}_{\Al}}(\Delta \vectorvar{p})$ forms with the positive $a_1$ axis in the subspace $\mathcal{S}_{\Al}$, which we denote by $\varphi(\mathbf{x}) \in [0, 2\pi)$.

Now, for any neural network function $f_{\Theta}: \mathbb{R}^6 \rightarrow \mathbb{R}^2$ with parameters $\Theta$, we will approximate the downwash forces $\mathbf{f}_{\text{ext}} $ felt by \Bravo{} as $F_{\Theta}: \mathbb{R}^9 \rightarrow \mathbb{R}^3$:
{\small
\begin{align}\label{equivariantmodel}
  F_{\Theta}(\mathbf{x}) \! = \! (R_{\mathcal{M}}^{\Al})^{\top} \! \bigg(\! \left[f_{\Theta}(h(\mathbf{x}))\right]_1 \! \cdot \![c(\varphi(\mathbf{x})), s(\varphi(\mathbf{x}))], \left[f_{\Theta}(h(\mathbf{x}))\right]_2 \!\bigg). 
\end{align}
}

\subsection{Proof of Equivariance}
\begin{theorem}\label{thm:equivariance}
The model $F_{\Theta}(\mathbf{x})$ proposed in (\ref{equivariantmodel}) for $\fext(\mathbf{x})$ satisfies Assumption \ref{assumption:equivariance}.
\end{theorem}

\proof
By Definition \ref{def:equivariance} of equivariance, we need to prove that for each $H \in H_{\Al}$,
\begin{align*}
    H \star_2 F_{\Theta}(\mathbf{x}) = F_{\Theta}(H \star_1 \mathbf{x}),
\end{align*}
where $\star_1$ and $\star_2$ are the group actions in Assumption \ref{assumption:equivariance}. 

First, we point out the fact that
\begin{align}\label{pf:S02}
\begingroup
\setlength\arraycolsep{2pt}
    H_{\Al} = \left\{ (R_{\mathcal{M}}^{\Al})^{\top} \begin{pmatrix}
    c(\omega) & -s(\omega) & 0\\
    s(\omega) & c(\omega) & 0\\
    0 & 0 & 1
    \end{pmatrix}R_{\mathcal{M}}^{\Al} \ \bigg| \ \omega \in [0, 2\pi)
    \right\}.
\endgroup
\end{align}
In other words, each $H \in H_{\Al}$ can be parameterized by the angle of rotation $\omega \in [0, 2\pi)$ about the axis $a_3$.
Let $H = (R_{\mathcal{M}}^{\Al})^{\top} \Omega R_{\mathcal{M}}^{\Al}$, where $\Omega$ is the rotation matrix in \eqref{pf:S02}.

As we previously discussed, we will first show that the feature mapping \eqref{featuremap} is \textit{invariant} to the action of $H_{\Al}$ on the input space, $\star_1$. 
Since $H$ is a rotation which fixes $a_3$, then
\begin{align*}
    & \left[ R_{\mathcal{M}}^{\Al} H \vectorvar{w} \right]_3 = \left[ \Omega R_{\mathcal{M}}^{\Al} \vectorvar{w} \right]_3 = \left[ R_{\mathcal{M}}^{\Al} \vectorvar{w} \right]_3, \ \forall \vectorvar{w} \in \mathbb{R}^3.
\end{align*}
Also, notice that 
\begin{align*}
   R_{\mathcal{M}}^{\Al}\proj{\mathcal{S}_{\Al}}(\mathbf{w}) = \begin{pmatrix}
    \mathbb{I}_{2 \times 2} & 0\\
    0 & 0
    \end{pmatrix} R_{\mathcal{M}}^{\Al}\mathbf{w}.
\end{align*}

But for any vector $\mathbf{w} \in \mathbb{R}^3$,
\begin{align*}
    R_{\mathcal{M}}^{\Al}\proj{\mathcal{S}_{\Al}}(H\mathbf{w}) &=
    \begin{pmatrix}
    \mathbb{I}_{2 \times 2} & 0\\
    0 & 0
    \end{pmatrix}R_{\mathcal{M}}^{\Al}H\mathbf{w}\\
    &= \Omega R_{\mathcal{M}}^{\Al} \proj{\Al}(\mathbf{w}).
\end{align*}
Since $\Omega$ and $R_{\mathcal{M}}^{\Al}$ are unitary, and the norm and dot product are preserved under unitary transformations, the previous two results imply $h(H \star_1 \mathbf{x}) = h(\mathbf{x})$.

Hence, it only remains to consider the polar angle that $\proj{\mathcal{S}_{\Al}}(H \Delta \mathbf{p})$, or equivalently $R_{\mathcal{M}}^{\Al} \proj{\mathcal{S}_{\Al}}(H \Delta \mathbf{p})$, forms with the positive $a_1$ axis in $S_{\Al}$. 
But $R_{\mathcal{M}}^{\Al} \proj{\Al}(H \Delta \mathbf{p}) = \Omega R_{\mathcal{M}}^{\Al} \proj{\Al}(\Delta \mathbf{p})$ is just $R_{\mathcal{M}}^{\Al} \proj{\Al}(\Delta \mathbf{p})$ rotated by $\omega$. 
Therefore, we know that
\begin{align*}
    \varphi(H \star_1 \mathbf{x}) \equiv \varphi(\mathbf{x}) + \omega \quad (\text{mod $2\pi$}).
\end{align*}
Let $\Omega_{2 \times 2} \in \mathbb{R}^{2 \times 2}$ be the submatrix formed by the first two rows and columns of $\Omega$. 
Using our established congruence,
\begin{align*}
    \Omega_{2 \times 2}[c(\varphi(\mathbf{x})), s(\varphi(\mathbf{x}))] &= [c(\varphi(\mathbf{x}) + \omega), s(\varphi(\mathbf{x}) + \omega)]\\
    &= [c(\varphi(H \star_1 \mathbf{x})), s(\varphi(H \star_1 \mathbf{x}))].
\end{align*}

Altogether, we conclude that
\begin{align*}
    &F_{\Theta}(H \star_1 \mathbf{x})\\
    =& (R_{\mathcal{M}}^{\Al})^{\top} \! \bigg( \! \!  [f_{\Theta}(h(\mathbf{x}))]_1 \Omega_{2 \times 2} [c(\varphi(\mathbf{x})), s(\varphi(\mathbf{x}))], [f_{\Theta}(h(\mathbf{x}))]_2 \! \! \bigg)\\
    =& (R_{\mathcal{M}}^{\Al})^{\top} \Omega\bigg(\! [f_{\Theta}(h(\mathbf{x}))]_1 \cdot [c(\varphi(\mathbf{x})), s(\varphi(\mathbf{x}))], [f_{\Theta}(h(\mathbf{x}))]_2 \! \bigg)\\
    =& H(R_{\mathcal{M}}^{\Al})^{\top} \! \bigg(\! [f_{\Theta}(h(\mathbf{x}))]_1 \cdot [c(\varphi(\mathbf{x})), s(\varphi(\mathbf{x}))], [f_{\Theta}(h(\mathbf{x}))]_2 \! \bigg)\\
    =& H \star_2 F_{\Theta}(\mathbf{x}).\qed
\end{align*}

\subsection{Shallow Learning}\label{sec:learning}
For our training pipeline, we collect time-stamped state and control information from real-world flights with \Alpha{} and \Bravo{}, and compute the input data points $\vectorvar{x}$ offline.
The labels that our model \eqref{equivariantmodel} learns to approximate are obtained from the feedback control equation \eqref{eq:rigid-body-lin}, $\fext = \vectorvar{a} - \bsym{u}(t).$

We choose the neural network $f_{\Theta}$ in \eqref{eq:rigid-body-lin} to be a shallow network with a single nonlinear activation
\begin{align}\label{shallownetwork}
    f_{\Theta}(\vectorvar{w}) = W^{(2)}\sigma\left(W^{(1)}\vectorvar{w} + b^{(1)} \right) + b^{(2)}, \quad \vectorvar{w} \in \mathbb{R}^6,
\end{align}
where $W^{(1)} \in \mathbb{R}^{32 \times 6}$, $b^{(1)} \in \mathbb{R}^{32}$, $W^{(2)} \in \mathbb{R}^{2 \times 32}$, $b^{(2)} \in \mathbb{R}^{2}$ are the network parameters $\Theta$ and $\sigma(\cdot) = \max(\cdot, 0)$ is the element-wise ReLU nonlinearity. 
$f_{\Theta}$ is trained to minimize the mean-squared error between the force prediction and label along all three inertial axes $\{\hat{e}_1, \hat{e}_2, \hat{e}_3\}$.

We justify our choice of a shallow neural network architecture via the $\SO{2}$ invariant feature mapping \eqref{featuremap}. 
For a neural network trained only on the raw input data $\mathbf{x}$, the model itself would be responsible for determining the geometries present in $\vectorvar{f}_{\text{ext}}$. 
However, for the equivariant model $F_{\Theta}$, the feature mapping \eqref{featuremap} encodes these geometries explicitly. 

Since our equivariant model is not responsible for learning the geometry of $\vectorvar{f}_{\text{ext}}$, we can reduce the complexity of $f_{\Theta}$ without sacrificing validation performance. 
We verify this claim empirically in Section \ref{subsec:modeltraining}.

\begin{figure}[t!]
    \centering
    \begin{subfigure}[t]{0.48\textwidth}
        \centering
        \includegraphics[width=\textwidth]
            {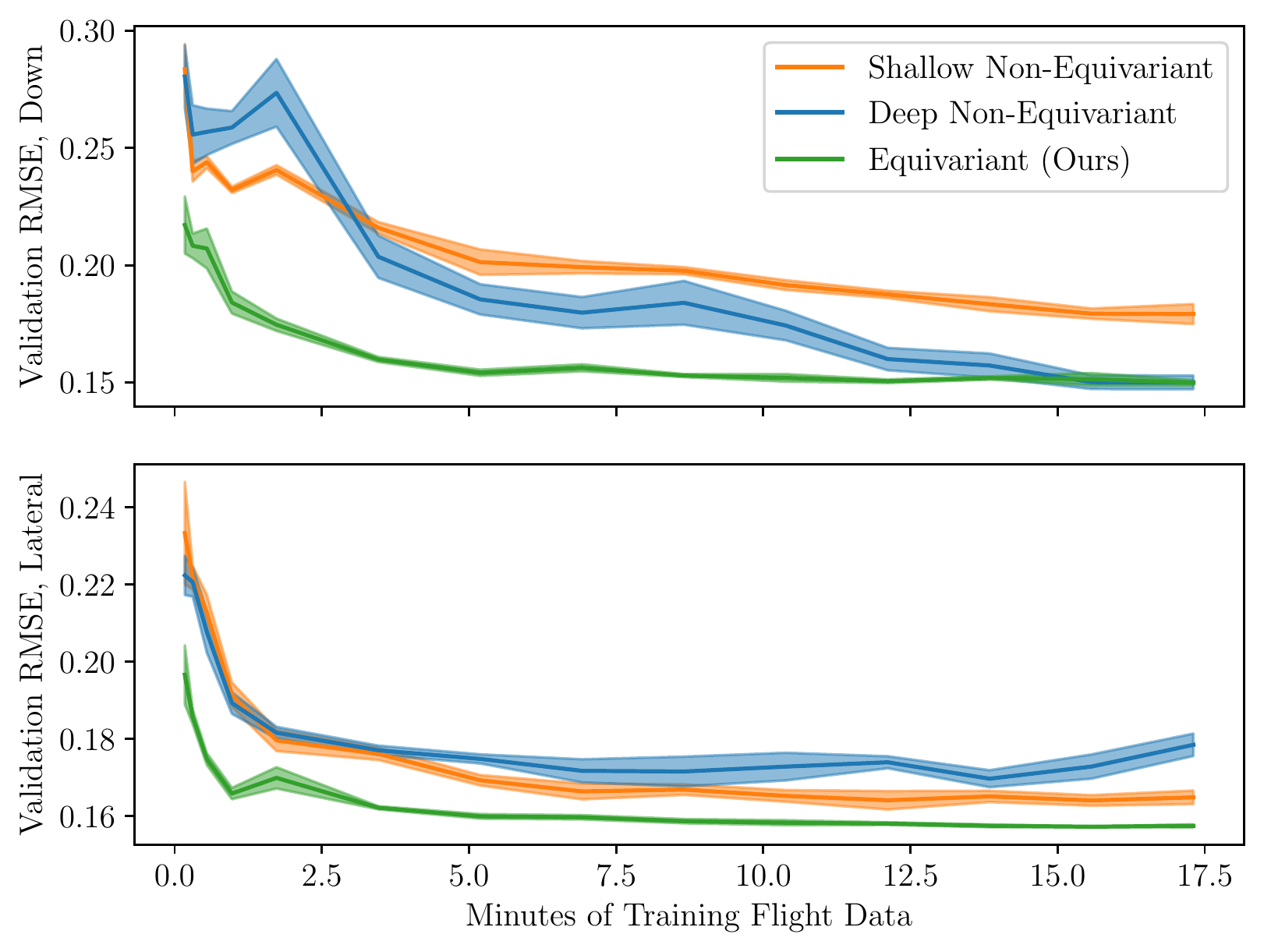}
        \vspace{-10pt}
    \end{subfigure} 
    
    \begin{subfigure}[t]{0.48\textwidth}
        \resizebox{\textwidth}{!}{%
        \begin{tabular}{|c|c|c|c|}%
        \hline
        \textbf{Model}  & \textbf{Shallow N-Equiv} & \textbf{Deep N-Equiv} & \textbf{Equivariant (Ours)} \\ \hline
        Number of Parameters      & $323$       & $9583$        & $258$          \\ \cline{1-1}
        Validation RMSE (5 min) & $0.263  \pm 0.005$    & $0.255  \pm 0.005$   & $\mathbf{0.222  \pm 0.001}$                  \\ \cline{1-1}
        Validation RMSE (15 min) &   $0.243  \pm 0.002$ & $0.229  \pm 0.003$   & $\mathbf{0.218  \pm 0.002}$                 \\ \cline{1-1}
        Position Error (lateral) [\SI{}{m}] & 0.126      & 0.125   & \textbf{0.104}     \\ \cline{1-1}
        Position Error (3D)  [\SI{}{m}]    & 0.132      & 0.127     & \textbf{0.106}       \\ \cline{1-1}
        Velocity Error (lateral) [\SI{}{m/s}] & 0.088      & 0.089     & \textbf{0.080}       \\ \cline{1-1}
        Velocity Error (3D)   [\SI{}{m/s}]   & 0.118      & 0.095     & \textbf{0.090}     \\ \hline
        \end{tabular}
        }
    \end{subfigure}
    \caption{\textit{Sample Efficiency and Accuracy.} Top: A visualization of the validation RMSE of the equivariant and non-equivariant models as a function of the training flight time. 
    For each training time, we compute the average validation RMSE across $5$ trials. 
    Bottom: Summary statistics for the equivariant and non-equivariant models. 
    Position and velocity tracking errors are reported for models trained on the full training dataset.}\label{fig:sampleefficiency+results}
\end{figure}
\section{Real-world Flight Experiments}\label{sec:experiments}
\begin{figure}[t!]
    \centering
    \includegraphics[width=0.48\textwidth]{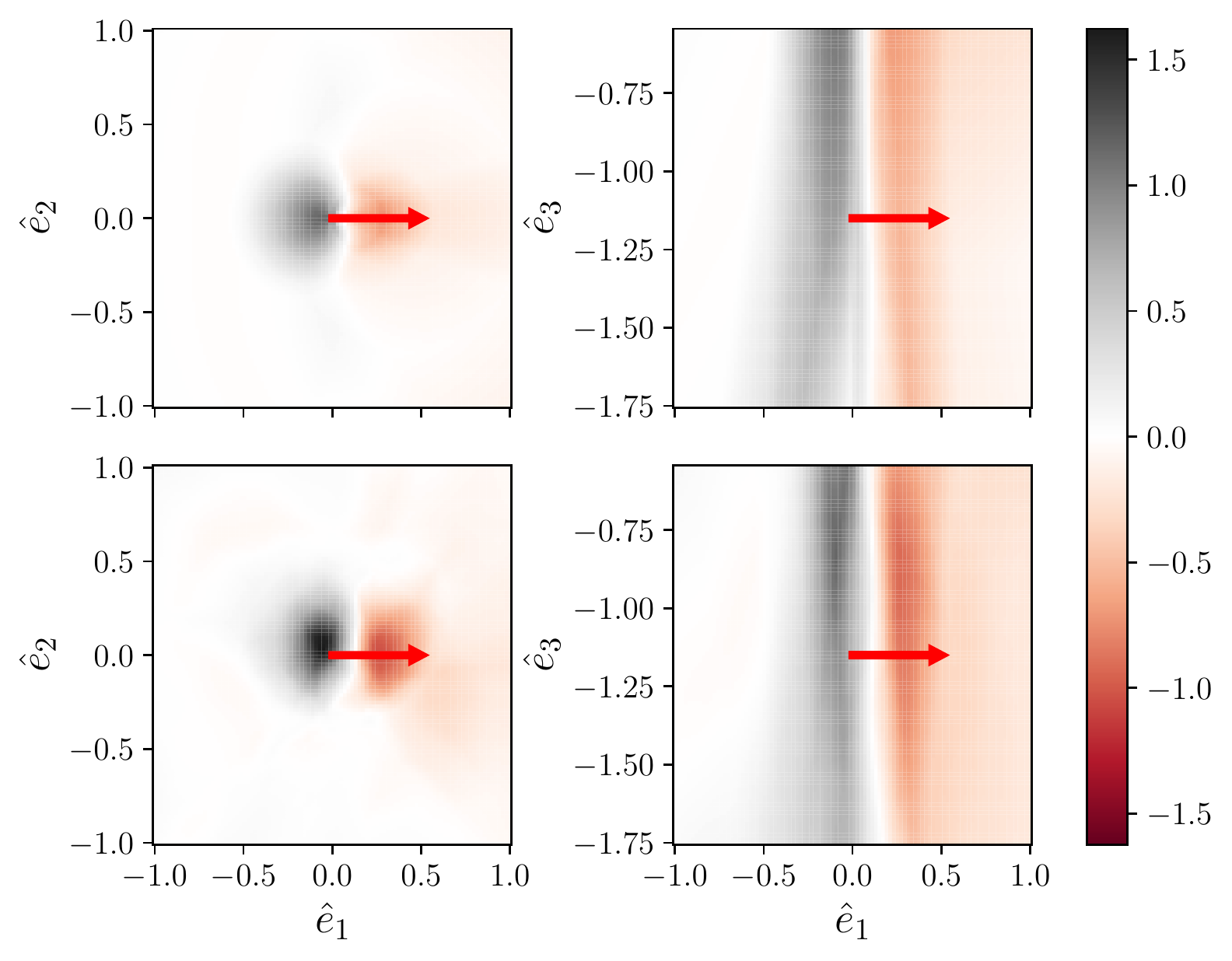}
    \caption{\textit{Downward Force Predictions.} Downward force predictions [\SI{}{m/s^2}] made by the equivariant model (top) and deep non-equivariant model (bottom).
    On the left (top-down view), \Alpha{} is hovering \SI{1}{m} above \Bravo{} at $(\hat{e}_1, \hat{e}_2) = (0,0)$.
    On the right (sagittal view), \Alpha{} is hovering \SI{0.1}{m} east of \Bravo{} at $(\hat{e}_1, \hat{e}_3) = (0,0)$.
    In each plot, \Bravo{} is moving with velocity $\vectorvar{v}^{\Br} = [0.5, 0, 0]^\top$.
    }
    \vskip-3ex
    \label{fig:force_pred_d}
\end{figure}

We conduct studies with our training procedure and present evaluations from real-world flight experiments.
For these, we will consider a special case of our model (\ref{equivariantmodel}) with $R_{\mathcal{M}}^{\Al} = \iden{3}{3}$, i.e., when \Alpha{}'s frame is a translation of the inertial frame and its instantaneous state is close to hovering.
We operate in an environment where other common external factors (such as wind) do not affect the vehicles.

\begin{figure}[tp]
\centering
  \begin{subfigure}[t]{0.48\textwidth}
    \centering
    \includegraphics[width=0.95\textwidth,trim={0 0 0 6.7cm},clip]
        {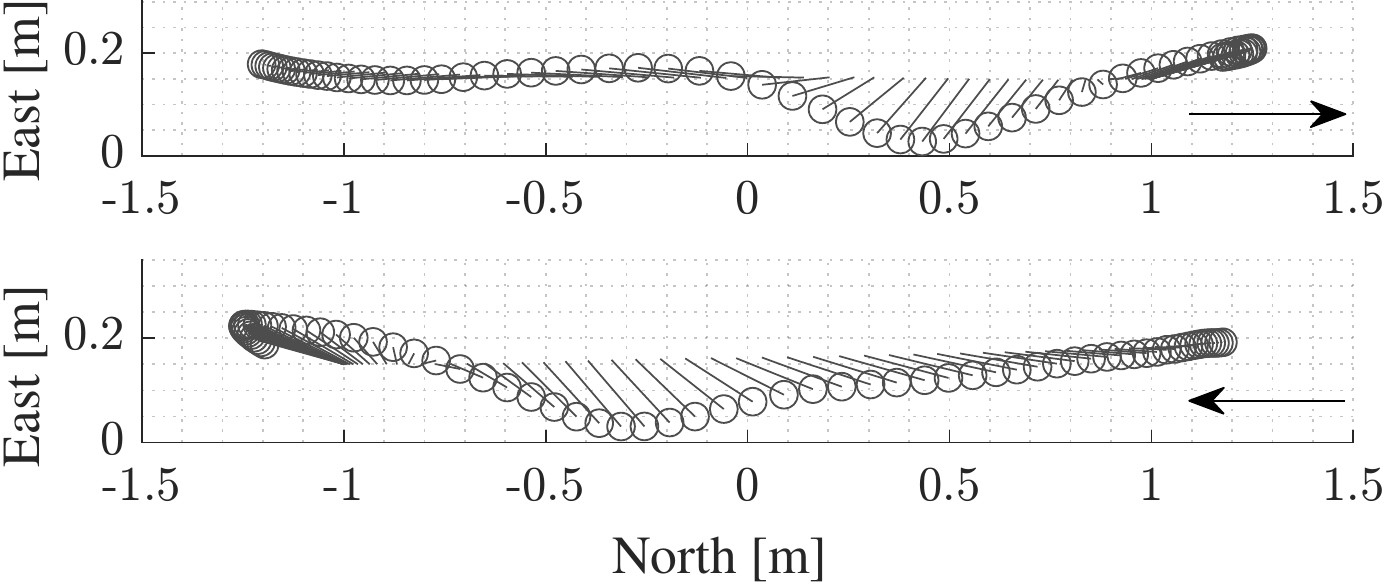}
    \caption{Position tracking errors (solid lines) for \Bravo{} when translating under \Alpha{}. The arrows indicate the direction of travel.}
    \label{fig:translation-trajs}
  \end{subfigure}
  \begin{subfigure}[t]{0.48\textwidth}
    \centering
    \includegraphics[width=\textwidth]{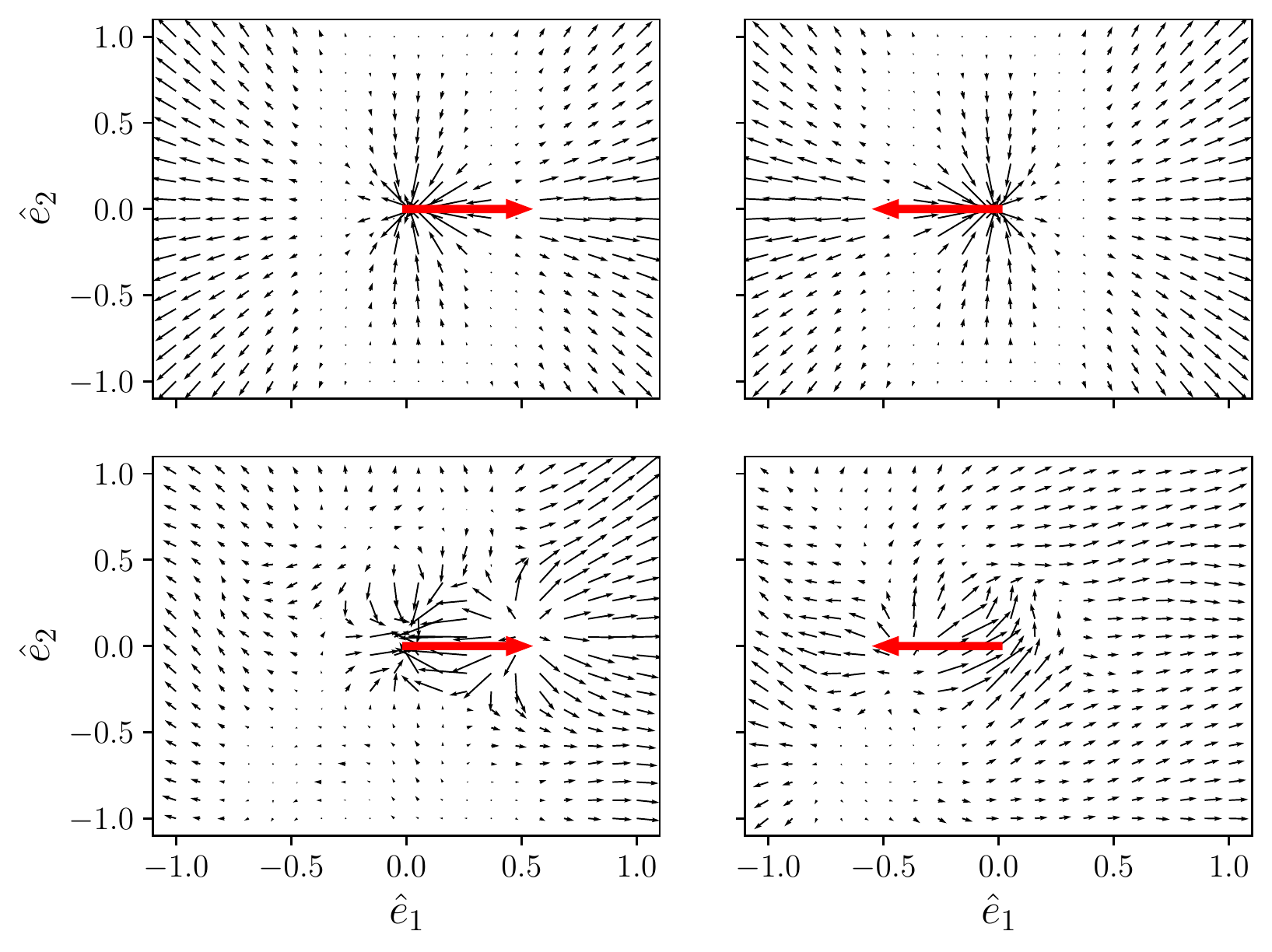}
    \caption{
    Lateral force predictions made by the equivariant model (top) and deep non-equivariant model (bottom). On the left, \Bravo{} is moving with velocity $\vectorvar{v}^{\Br} = [0.5, 0, 0]^{\top}$, and on the right it is moving with velocity $-\vectorvar{v}^{\Br}$. Since $\SO{2}$ equivariance is not explicitly imposed in the non-equivariant model, its force predictions do not satisfy the assumed geometry.}
    \label{fig:quiver}
  \end{subfigure}
  \caption{\textit{Lateral Forces.} Force predictions and errors during a transition under \Alpha{} with a vertical separation of  \SI{0.8}{m}. \Alpha{} is hovering at $(\hat{e}_1, \hat{e}_2) = (0,0)$.}
  \label{fig:force-equiv}
\end{figure}

\subsection{Sequential Data Collection}\label{sec:data-collection}

In order to train our model, we first need to collect a dataset of real-world flights.
This is a difficult task, since without a compensation model \textit{a priori}, physical and control limitations will prevent the vehicles from flying in close proximity.
To solve this problem, we adopt a sequential approach similar to \cite{shi2020neural} that splits data collection into different `stages'.
In \textsf{stage-0}, we fly the vehicles with a relatively large vertical separation, $\SI{1.75}{m}$ to $\SI{1.35}{m}$, so that the forces acting on \Bravo{} can be compensated for by a disturbance-rejecting nominal controller.

This controller has no prior knowledge of the exogenic forces, so it simply relies on its feedback control input, $\bsym{u}(t)$, to track the desired reference trajectories according to (\ref{eq:rigid-body-lin}).
However, in regions of strong downwash forces, the measured accelerations, $\vectorvar{a}$, will not correspond to the desired acceleration input, $\bsym{u}$.
These deviations become the computed force labels (represented in mass-normalized acceleration units) for this stage.

After training a \textsf{stage-0} model using (\ref{equivariantmodel}), we deploy it online within the control loop such that $\bsym{u}_\vectorvar{f}$ now provides the feedback regulation.
As a result, we can now decrease the separation between the vehicles and obtain a new \textsf{stage-1} dataset along with its labels (the residual disturbances our  \textsf{stage-0} model is unable to account for).
The \textsf{stage-1} dataset is concatenated with the \textsf{stage-0} dataset and used to train the  \textsf{stage-1} model.

We repeat this process for a total of three stages, so that at \textsf{stage-2} the vehicles have a vertical separation of only $\approx\SI{0.5}{m}$ (approx.~two body-lengths).
Our full training dataset corresponds to approximately $17$ minutes of flight data. 
Note that we can always extract the correct force labels at \textsf{stage-i} by logging the model predictions $F_{\Theta}(\mathbf{x})$ from \textsf{stage-(i-1)} and subtracting these from the control.

\subsection{Study: Model Training}\label{subsec:modeltraining}
We first study the effect of geometric priors on the learning algorithm for modelling downwash forces $\fext(\mathbf{x})$.

\textbf{Non-equivariant Baselines.}
In order to analyze the effect of our geometric priors, we must first introduce two models to which we can compare our $\SO{2}$-equivariant model \eqref{equivariantmodel}.
These ``non-equivariant" models should not exploit the known geometry of $\fext$ delineated in Assumption \ref{assumption:equivariance}. 

The first non-equivariant model we propose has the same architecture as the shallow neural network \eqref{shallownetwork}, with the exception that the input to the network is $[\Delta \vectorvar{p}, \vectorvar{v}^{\Br}] \in \mathbb{R}^6$ rather than invariant feature representation $h(\vectorvar{x})$.
Note that $\vectorvar{v}^{\Al}$ is not included because of the near-hover assumption that we specified at the beginning of the section.

We also compare our equivariant model against the SOTA eight-layer, non-equivariant neural network discussed in Section \ref{sec:introduction} \cite{shi2020neural, shi2021neural2}. 
During training, we bound the singular values of the weight matrices to be at most $2$. 
This normalization technique, called ``spectral normalization," constrains the Lipschitz constant of the neural network \cite{shi2020neural, shi2021neural2}.

\textbf{Efficiency of Geometric Learning.}
As we suggested in Section \ref{sec:introduction}, the primary benefit of imposing geometric priors on a learning algorithm is that they have been empirically shown to improve sample efficiency \cite{wang2023surprising, wang2022equivariant}.

We investigate the sample efficiency of our $\SO{2}$-equivariant downwash model by considering the validation root mean-squared error (RMSE) as a function of the length of our training flights. 
We shorten the full training dataset by shortening each stage of data collection proportionally (ex.~a total training time of three minutes corresponds to one minute of flight for each stage).
Our validation dataset is roughly equal in size to the full training dataset.
 
In Figure \ref{fig:sampleefficiency+results}, we observe that although the validation loss of the shallow non-equivariant network plateaus after approximately $10$ minutes of training flight data, it cannot represent the downward aerodynamic forces as accurately as the other models (i.e.~greater bias).
Conversely, while the deep non-equivariant network accurately learns the downward forces, it requires much more training data to do so (i.e.~lower sample efficiency).
Neither non-equivariant model learns the lateral forces as accurately as the equivariant model. 

Our equivariant model \eqref{equivariantmodel}, on the other hand, displays both high sample efficiency and low bias. 
With only $5$ minutes of flight data, it learns the lateral and downward forces more accurately than both non-equivariant models do with $15$ minutes of data.

\textbf{Visualizing Downwash Predictions.}
In Figure \ref{fig:force_pred_d}, we visualize the force predictions that our equivariant model $F_{\Theta}(\mathbf{x})$ makes in the $\hat{e}_3$ direction. 
When \Bravo{} passes through the downwash region of \Alpha{}, there is a highly repeatable pattern in which it is first subjected to a positive force, which pushes it towards the ground, followed by a negative force, which pulls it upwards.
The magnitudes of these positive and negative forces are dependent upon \textit{(i)}  \Bravo{}'s speed as it passes through the downwash region, and, \textit{(ii)} its distance from \Alpha{} in both $\mathcal{S}_{\Al}$ and  $\mathcal{S}_{\Al}^{\perp}$.
Similar patterns have been documented by previous downwash models \cite{shi2020neural}.

From Figure \ref{fig:quiver}, we observe that $F_{\Theta}$ also uncovers consistent patterns in the lateral axes $\hat{e}_1$ and $\hat{e}_2$.
When \Bravo{} translates laterally underneath \Alpha{}, it is first pushed radially outwards, then pulled inwards immediately upon passing under \Alpha{}, and lastly, pushed radially outwards once it has passed \Alpha{}.
These inwards forces are strongest when \Bravo{} is traveling at a high speed.
In Figure \ref{fig:translation-trajs}, we show that the model's predictions are consistent with the observed deviations of \Bravo{} from its trajectory.

We believe that we are the first to demonstrate these lateral force patterns via a machine learning approach. 

\subsection{Real-World Experiments}
\label{subsec:realworld-exp}

We evaluate the performance of our trained equivariant model \eqref{equivariantmodel} in two real-world experiments, and contrast it against a baseline controller as well as the deep non-equivariant model. Each model is trained on the full training dataset as described in Section \ref{sec:data-collection}.

Our tests use two identical quadrotor platforms that are custom-built using commercial off-the-shelf parts.
These span $\SI{0.26}{m}$ on the longest body-diagonal,
and weigh $\approx\SI{0.7}{kg}$ (including batteries).
Each platform is equipped with a Raspberry Pi 4B (8GB memory) on which we run our control, estimation and model evaluations.
The model-based LQG flight control and estimation \eqref{eq:rigid-body-lin} is performed by \textit{Freyja}~\cite{shankar2021freyja}, 
while the neural-network encapsulation is done through \textsf{PyTorch}.
Our systems run in a decentralized mode, with independent positioning data obtained through an OptiTrack motion-capture system.
The controller and the model iterations are performed at \SI{50}{Hz} and \SI{45}{Hz} respectively.

Prior to conducting tests with \Bravo{} in motion, we first ensure that a stationary hover under \Alpha{} is stable when the model's predictions are incorporated into \Bravo{}'s control loop.
This is essential to validate empirically that the predictions made by the model do not induce unbounded oscillations on \Bravo{}.
The table in Figure \ref{fig:sampleefficiency+results} lists the quantitative results averaged across all our experiments, compared against baseline methods.

\begin{figure}
    \centering
    \includegraphics[width=0.48\textwidth]
        {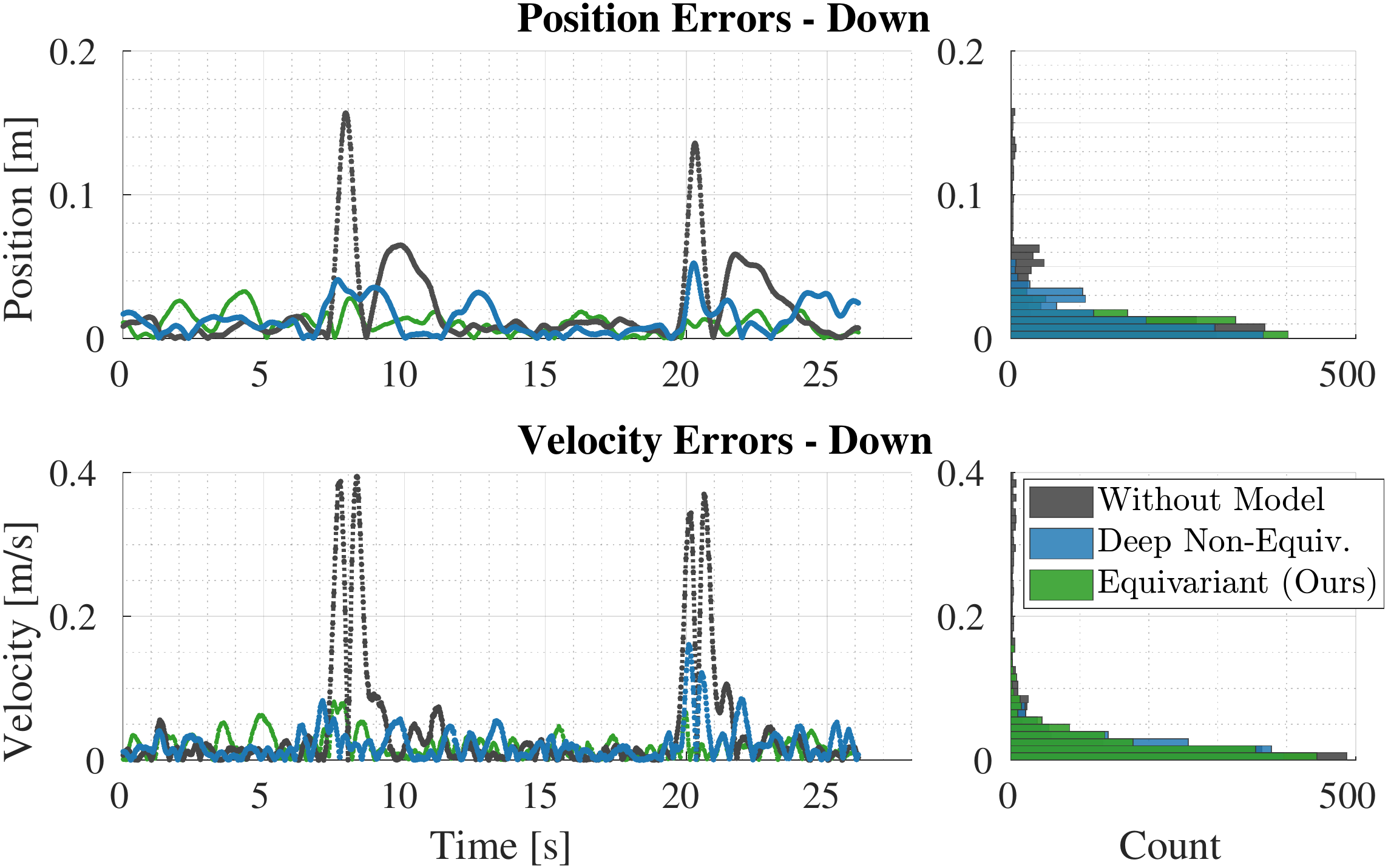}
    \\[1ex]
    \includegraphics[width=0.48\textwidth]
        {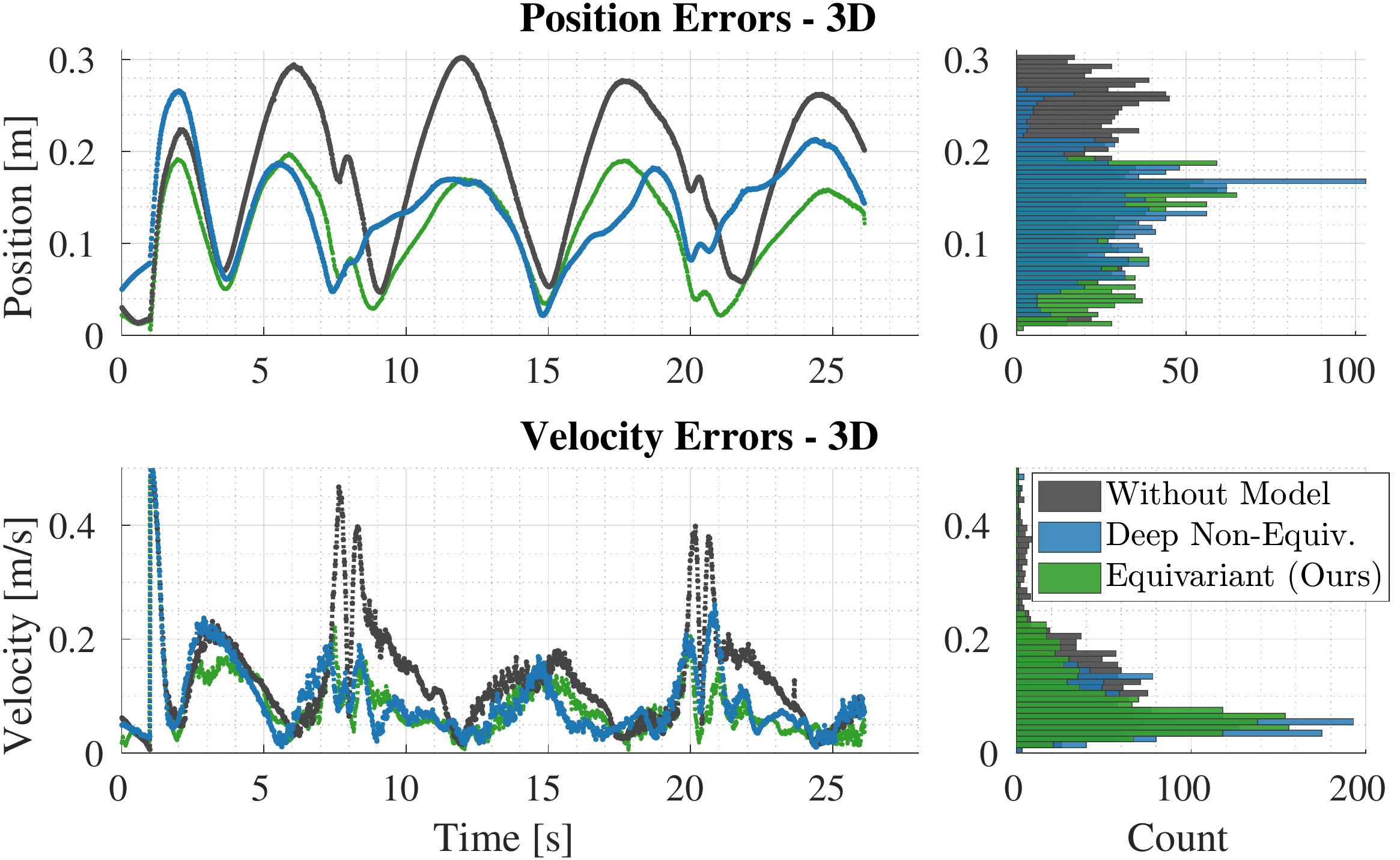}
    \caption{\textit{Lemniscate.}
    An evaluation of \Bravo{}'s trajectory tracking performance with \Alpha{} hovering at $\hat{e}_3 = \SI{-2.5}{m}$. 
    The first two rows show the evolution of the position and velocity tracking errors, as well as their distributions. 
    The last two rows show the same statistics in the 3D space. 
    Our model improves the position and velocity error distributions when compared against a baseline controller (without any model) and the non-equivariant model.}
    \vspace{-10pt}
    \label{fig:lemniscate-results}
\end{figure}

\begin{figure}
    \centering
    \includegraphics[width=0.48\textwidth]
        {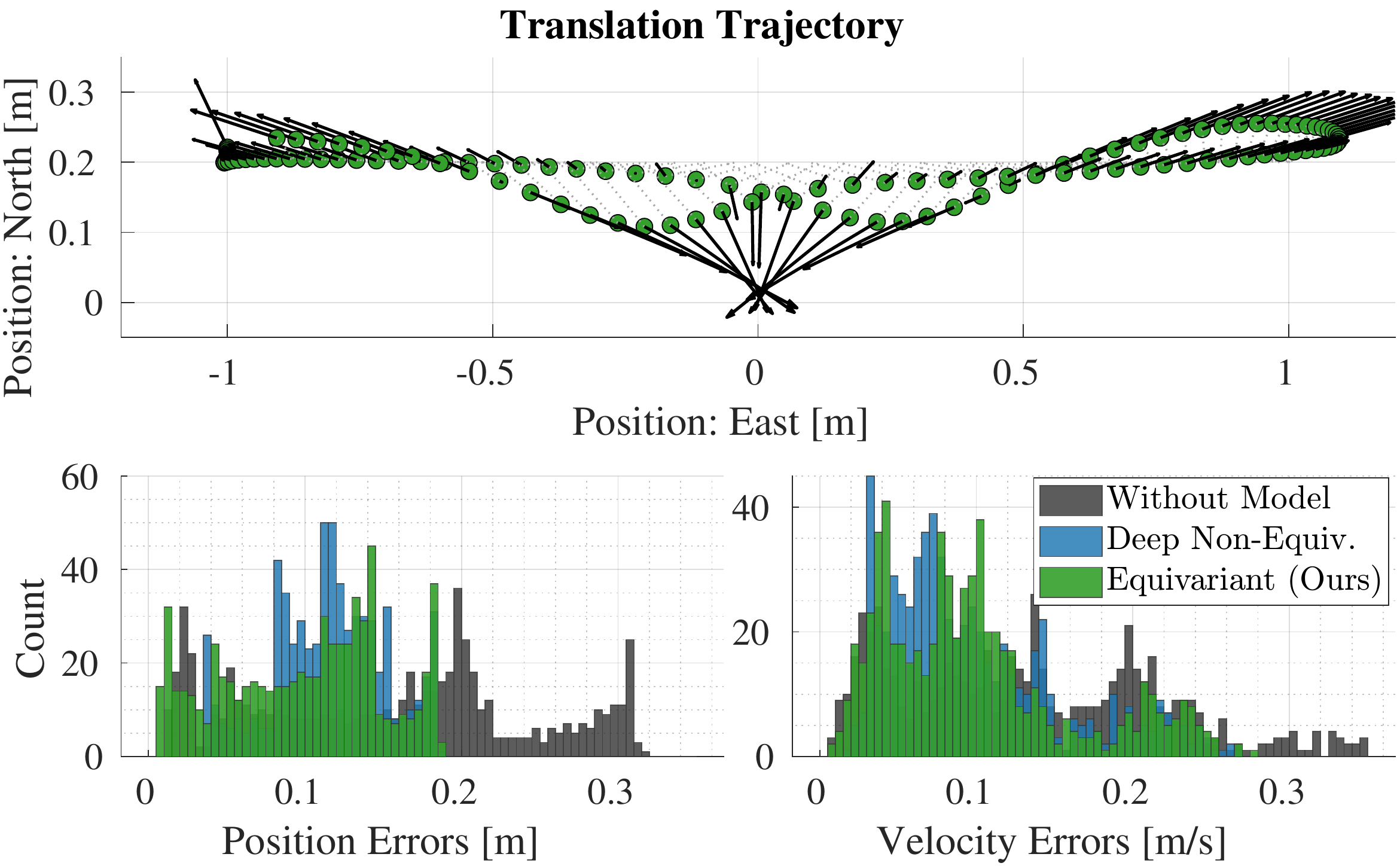}
    \caption{\textit{Translation.}
    An evaluation of \Bravo{}'s flight performance over a lateral transect beneath \Alpha{}.
    This trajectory is replicated from Figure~\ref{fig:translation-trajs}. Solid black arrows represent the force predictions made by the equivariant model.
    Applying these predictions, we reduce mean position and velocity tracking errors by almost \SI{36}{\percent} (see Figure \ref{fig:sampleefficiency+results}), with over \SI{55}{\percent} improvement in $\hat{e}_3$ axis alone.
    }
    \vspace{-5pt}
    \label{fig:transect-results}
\end{figure}

\textbf{Lemniscate Trajectory.}
We now evaluate the model deployed in a more dynamic scenario where \Bravo{} is commanded to follow a lemniscate trajectory (`figure-eight') under \Alpha{}.
This exposes the model to many different regions of the state space, while also requiring \Bravo{} to make continuous changes to its accelerations.

Figure~\ref{fig:lemniscate-results} shows tracking results from executing one complete  period of this trajectory.
We observe that deploying our model produces a significant shift in the distribution of both position and velocity errors.
Without any model, \Bravo{} loses vertical position tracking (top row) twice as it makes the two passes directly beneath \Alpha{}, seen near \SI{8}{s} and \SI{20}{s}.
These spikes, also noticeable as vertical velocity errors (second row), are ``absorbed" due to the predictions made by our equivariant model (as well as the baseline deep non-equivariant model).
The equivariant model produces an improvement of nearly \SI{51}{\percent} in both position and velocity tracking, whereas the non-equivariant model is still able to provide almost \SI{36}{\percent} and \SI{46}{\percent} improvement, respectively.

Our model's ability to represent geometric patterns in the lateral plane is also apparent when considering the full 3D errors (third and fourth rows).
The non-equivariant model already improves position tracking by \SI{24}{\percent} (\SI{0.137}{m} from \SI{0.181}{m}) and velocity tracking by nearly \SI{28}{\percent} (\SI{0.091}{m/s} from \SI{0.128}{m/s}).
The equivariant model decreases position errors further (down to \SI{0.113}{m}, \SI{37}{\percent} improvement), and also reduces velocity tracking errors (down to \SI{0.081}{m/s}, a \SI{36}{\percent} improvement).

\textbf{Translation Trajectory}
Next, we perform an analysis of \Bravo{}'s tracking performance and the model's responses while executing a horizontal transect under \Alpha{}.
This trajectory is the same as the one shown in Figure~\ref{fig:translation-trajs}, and is useful because it drives \Bravo{} rapidly through regions of near-zero to peak disturbances.

Figure \ref{fig:transect-results} illustrates key results from one back-and-forth trajectory parallel to the $\hat{e}_2$ axis.  $\hat{e}_1 = 0.2$ is fixed, and \Bravo{} is at a fixed vertical separation of \SI{0.6}{m} with \Alpha{} hovering at $ \mathbf{p}^{\Al} = [0, 0, -2.5]$.
The first row shows the actual trajectory executed by \Bravo{} with our equivariant model deployed (green circles), with an overlay of the force predictions made by the model (solid black arrows).
We first point out that the pattern is similar to the one found in Figure \ref{fig:translation-trajs}, but the peak errors have decreased significantly.

The distributions of errors shown in the second row demonstrate that the magnitudes of these predictions are also justified.
Even though \Bravo{} is not directly underneath \Alpha{} in these tests, it is still well within \Alpha{}'s downwash region.
Across experiments, we observe a reduction in the mean 3D positioning error to \SI{0.098}{m} (from \SI{0.154}{m}), corresponding to an improvement of almost \SI{36}{\percent} (the peak error is also reduced similarly).
Velocity tracking error also shows a similar trend, with an average improvement of \SI{34}{\percent}.
Considering only the vertical tracking performance in these tests (not shown in figures), these statistics jump to \SI{55}{\percent} and \SI{49}{\percent} (for position and velocity, respectively).
\section{Conclusion}
This article proposes a sample-efficient learning-based approach for modelling the downwash forces produced by a multirotor on another.
In comparison to previous learning-based approaches that have tackled this problem, we make the additional assumption that the downwash function is rotationally equivariant about the vertical axis of the leader vehicle. 
Through a number of real-world experiments, we demonstrate that the equivariant model outperforms baseline feedback control as well as SOTA learning-based approaches. The advantage of our equivariant model is greatest in regimes where training data is limited. 

In the future, we will further explore the potential of our equivariant model through flight regimes with larger force magnitudes. 
This includes outdoor flights, where the leader and follower can move at sustained greater speeds, and interact with ambient wind.
Finally, we will consider how to model the geometries present in a multi-vehicle system. 
One na\"ive approach to modelling multi-vehicle downwash would be to employ our two-vehicle model and sum the individual force contributions produced by each multirotor in the system. 
However, it may be the case that  individual downwash fields interact in highly nonlinear ways, in which case a more complex model of the multi-vehicle geometry would be necessary.

\section*{Acknowledgment}
We thank Heedo Woo for his contributions to the construction of the quadrotors, and Wolfgang H\"onig for his clarifications about the sequential data collection in \cite{shi2020neural}.

\bibliographystyle{IEEEtran.bst}
\bibliography{bibliography.bib, aj_refs}

\end{document}